\documentclass[runningheads]{llncs}
\usepackage[T1]{fontenc}
\usepackage{graphicx}
\usepackage{todonotes}
\usepackage{enumitem}
\usepackage{amsmath}
\begin{document}
\title{UltraRay: Introducing Full-Path Ray Tracing in Physics-Based Ultrasound Simulation}
\titlerunning{UltraRay}

\author{Felix Duelmer \inst{1,2,3,4}\thanks{Corresponding Author} \and
Mohammad Farid Azampour \inst{1,2}, Magdalena Wysocki \inst{1, 2} \and Nassir Navab\inst{1, 2}}
\authorrunning{F. Duelmer et al.}
%

%
\institute{
Chair for Computer-Aided Medical Procedures and Augmented Reality, School of Computation, Information and Technology, Technical University of Munich, Munich, Germany \and
Munich Center for Machine Learning (MCML), Munich, Germany \and
Chair of Biological Imaging, Central Institute for Translational Cancer Research (TranslaTUM), School of Medicine and Health, Technical University of Munich, Munich, Germany \and
Institute of Biological and Medical Imaging, Bioengineering Center, Helmholtz Zentrum München, Neuherberg, Germany
}
\maketitle              
\begin{abstract}

Traditional ultrasound simulators solve the wave equation to model pressure distribution fields, achieving high accuracy but requiring significant computational time and resources. To address this, ray tracing approaches have been introduced, modeling wave propagation as rays interacting with boundaries and scatterers. However, existing models simplify ray propagation, generating echoes at interaction points without considering return paths to the sensor. This can result in unrealistic artifacts and necessitates careful scene tuning for plausible results.

We propose a novel ultrasound simulation pipeline that utilizes a ray tracing algorithm to generate echo data, tracing each ray from the transducer through the scene and back to the sensor. To replicate advanced ultrasound imaging, we introduce a ray emission scheme optimized for plane wave imaging, incorporating delay and steering capabilities. Furthermore, we integrate a standard signal processing pipeline to simulate end-to-end ultrasound image formation.

We showcase the efficacy of the proposed pipeline by modeling synthetic scenes featuring highly reflective objects, such as bones. In doing so, our proposed approach, UltraRay, not only enhances the overall visual quality but also improves the realism of the simulated images by accurately capturing secondary reflections and reducing unnatural artifacts. By building on top of a differentiable framework, the proposed pipeline lays the groundwork for a fast and differentiable ultrasound simulation tool necessary for gradient-based optimization, enabling advanced ultrasound beamforming strategies, neural network integration, and accurate inverse scene reconstruction.

\keywords{Ray Tracing  \and Monte Carlo \and Ultrasound \and Simulation}
\end{abstract}
\section{Introduction}

Ultrasound simulation is important in medical imaging, with applications in areas like training sonographers \cite{dietrich2023ultrasound,lobo2024emerging}, designing and testing transducers \cite{canney2008acoustic,clement2000field,ghanem2018field}, or, more recently, generating data for training neural networks \cite{velikova2023lotus,behboodi2019ultrasound,amadou2024cardiac}. 

Traditional ultrasound simulators solve the wave equation using Green's function, either with linear methods \cite{jensen2002ultrasound} or nonlinear approaches \cite{treeby2010k}. While these methods are accurate, they are computationally intensive, as they require solving complex mathematical equations over fine spatial and temporal grids. This makes them slow and difficult to apply for real-time applications \cite{peng2019real}, large-scale simulations \cite{burman2024large}, or waveform inversion tasks \cite{lucka2021high}. To address these limitations, ray tracing algorithms were introduced \cite{wein2007simulation,shams2008real,kutter2009visualization}, initially for CT-US registration. Over time, significant advancements were made to improve ray tracing methods. Gao \textit{et al.}~\cite{gao2009fast} proposed a convolutional ray casting approach that efficiently simulates scatterers by convolving the point spread function (PSF) with scatterer distributions. Later developments improved tissue boundary interactions \cite{burger2012real,amadou2024cardiac}, exchanged ray casting with ray tracing \cite{burger2012real}, and incorporated Monte Carlo ray tracing (MCRT) for greater realism \cite{mattausch2018realistic,amadou2024cardiac}.

However, current ray tracing methods typically only account for the ray traveling from the transducer to the scattering or reflection event, directly adding an echo to the time-echo signal without verifying whether the ray can be received at the sensor. As a result, beamforming algorithms are generally excluded from these simulators, as they assume the data produced by the ray tracing algorithm already represents beamformed channel data. This simplification can lead to unrealistic reflections, introducing implausible artifacts and reducing the overall realism of the simulated images.

In the synthesis of natural images using ray tracing, similar challenges arise in the visible light domain as in ultrasound, including phenomena like reflections, refractions, and scattering. To address these, physically based renderers \cite{pharr2023physically} have been developed to create photorealistic images by explicitly simulating the complete light transport process using ray tracing. A growing research area focuses on the invertibility of this rendering process, enabling the reconstruction of scenes from acquired images. Efficient ray tracing simulators have been introduced to tackle these challenges \cite{li2018differentiable,nimier2019mitsuba,zhang2019differential}.

The invertibility of rendering processes has also gained attention in the ultrasound domain, where achieving differentiable, physically accurate, and fast simulations remains challenging due to the computational demands of traditional methods. Recent works, such as UltraNeRF \cite{wysocki2024ultra}, address this by utilizing a fast rendering process combined with implicit neural representations for efficient ultrasound image synthesis. However, simplifications in their underlying equations currently limit their ability to fully capture the complexity of ultrasound physics.

To address this gap, we propose a novel, fast ray tracing simulator for ultrasound built on a differentiable framework. The proposed simulator accounts for the complete path of sound waves through tissue and incorporates phase information, enhancing realism while mimicking the real ultrasound image formation process. Our key contributions are as follows:

\begin{itemize}[label=\textbullet]
    \item \textbf{Full Round-Trip Ray Tracing}: We introduce a method that traces rays from their emission at the transducer to their reception, modeling interactions with tissues using physically-based rendering equations.
    \item  \textbf{Versatile Echo Acquisition Strategies}: The proposed simulator supports various acquisition methods, such as traditional channel-based acquisition or synthetic aperture acquisition. We demonstrate this capability using plane-wave imaging, commonly employed in modern ultrasound systems. Due to the independence of the rays from each other, acquisitions that would require multiple transmission events in wave base simulators can be simulated in parallel in the proposed approach.
    \item \textbf{End-to-End Ultrasound Imaging}: We include a complete signal processing pipeline, implementing steps such as beamforming and log-compression, to simulate the complete imaging formation process in ultrasound.
    \item \textbf{Code Availability}: The simulator’s code and example scenes are openly accessible to support future research and development at \\ \url{https://github.com/Felixduelmer/UltraRay}.

\end{itemize}

\begin{figure}[hbt!]
    \centering
    \includegraphics[width=0.95\textwidth]{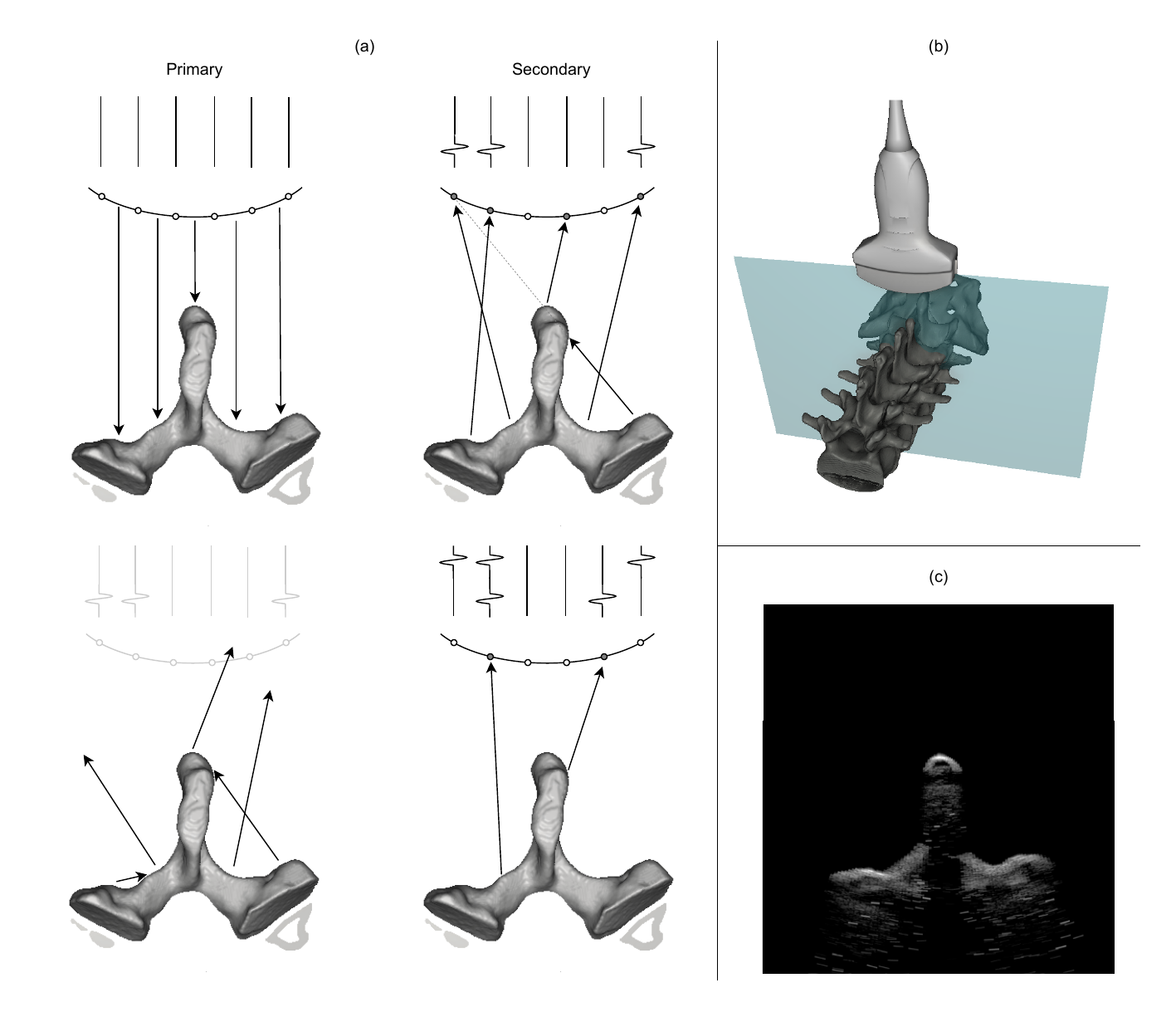}
    \caption{a) 
    Visualization of the Ray Tracing Process: The process begins with emitting a primary ray front from the transducer. At each interaction point with the vertebrae, secondary rays are cast toward the center of a randomly sampled transducer element. If a secondary ray is not blocked, its contribution is added to the corresponding time-pressure signal of that transducer element. The primary rays then continue their traversal by evaluating the surface interaction and determining a new direction. This process repeats—casting secondary rays at subsequent interaction points—until all primary rays become inactive, either by meeting a stopping condition or failing to intersect an object in the scene. In (b), we illustrate the virtual scene being imaged, which includes the transducer and acquisition plane. In (c), we present the resulting B-mode image generated by the proposed method.
    }
    \label{fig:model-arch}
\end{figure}

\section{Methodology}

We begin by introducing the underlying equations used in the simulator (Section \ref{sec:ray_tracing_for_utlrasound}) and explain their adaptation into a Monte Carlo ray tracing approach (Section \ref{sec:monte_carlo_ray_tracing}). Next, we present our proposed method, detailing the ray tracing process, including ray emission, traversal through the scene, and the transducer sampling strategy. Finally, we describe the subsequent signal processing pipeline (Section \ref{sec:proposed_approach}).

\subsection{Ray Tracing in Ultrasound}\label{sec:ray_tracing_for_utlrasound}

In ultrasound imaging, we aim to determine the pressure signal $P$ at a transducer element $e$ over time $t$, based on a specific acquisition pattern. These pressure signals per transducer element form the input to the standard signal processing pipeline, transforming the time signals into a B-mode image. By modeling the incoming pressure waves as pressure rays, the contribution to an individual transducer element can be expressed as:
\begin{equation}
    P(e, t) = \int_{\Omega} \int_{A} P_i(\mathbf{x}, t, \omega_i) f_d(\omega_i) d\omega \, da
\end{equation}

Here, $\omega_i$ represents the direction of an incoming ray over the hemisphere $\Omega$, originating from a point $\mathbf{x}$. We additionally take the integral over the surface $A$ of the transducer element plane. The term $P_i$ denotes the incoming pressure signal from that direction over time, while the directivity function $f_d$ weighs the contribution of pressure from each direction, accounting for the directional sensitivity of the transducer element.

To analyze the contributing pressure $P_i$ originating from a point $\mathbf{x}$, it is necessary to evaluate the incident, outgoing, and emitted rays at that location. To achieve this, we draw inspiration from the treatment of light interactions in physically based rendering systems and adapt those principles to ultrasound.

Physically based rendering computes light paths by solving the general rendering equation at a point $\mathbf{x}$, as follows:
    \begin{equation}
    L_o(\mathbf{x}, \omega_o) = L_e(\mathbf{x}, \omega_o) + \int_{\Omega} f_r(\mathbf{x}, \omega_i, \omega_o) L_i(\mathbf{x}, \omega_i) (\mathbf{n} \cdot \omega_i) \, d\omega_i
    \end{equation}
In this equation, $L_o$, $L_e$, and $L_i$ are the outgoing, emitted, and incident radiance at surface point $x$, respectively. The terms $\omega_o$ and $\omega_i$ denote the outgoing and incident directions, while $n$ is the surface normal at $\mathbf{x}$. The integrand computes the contribution of incoming light rays across the hemisphere $\Omega$, weighting each ray with a bidirectional scattering distribution function (BSDF), $f_r$, and a geometric term $(\mathbf{n} \cdot \omega_i)$. $\cdot$ represents the dot product. The BSDF models how light interacts with each material based on the properties and directions of incoming and outgoing rays.

For ultrasound imaging, we adapt this framework by replacing radiance with the pressure $P$ over time $t$, similarly to \cite{amadou2024cardiac}. The resulting equation becomes:
\begin{equation}
P_o(\mathbf{x}, t, \omega_o) = P_e(\mathbf{x}, t, \omega_o) + \int_{\Omega} f_r(\mathbf{x}, \omega_i, \omega_o) P_i(\mathbf{x}, \omega_i) (\mathbf{n} \cdot \omega_i) \, d\omega_i
\end{equation}
In contrast to \cite{amadou2024cardiac}, we explicitly include the emitter pressure function $P_e$ in the equation to account for the complete round-trip, encompassing both emission and reception of pressure waves at the transducer.

While it is theoretically possible to compute these integrals iteratively for all elements and points, this approach quickly becomes computationally prohibitive due to the immense workload. To address this challenge, we adopt a MCRT scheme, widely used in physically based rendering, to efficiently simulate and aggregate the contributions of rays in the ultrasound domain.

\subsection{Monte-Carlo Ray tracing} \label{sec:monte_carlo_ray_tracing}

In order to facilitate ray tracing while maintaining realism, we are adapting the strategy from \cite{amadou2024cardiac,mattausch2018realistic} to model ultrasound simulation as a MCRT scheme. Rewriting the pressure signal $P$ at transducer element $e$, results in:

\begin{equation}\label{eq:mc_emitter}
    P(e, t) =\frac{1}{N} \sum_{i=1}^N \frac{P_i(\mathbf{x}, t, \omega_i)f_d(\omega_i)}{p_t(\omega_i)}
\end{equation}

where the integral over the transducer surface and the hemisphere is replaced with a random sampling strategy. Each sampled direction $\omega_i$ contributes to the estimate, weighted by the inverse of its sampling probability $p_t(\omega_i)$. Here, $p_t$ is the transducer's probability density function (PDF), which describes the likelihood of sampling a ray in direction $\omega_i$. This PDF includes contributions from the transducer's directivity function $f_d(\omega_i)$, ensuring that directions are sampled in proportion to their importance based on the transducer's sensitivity. The result is normalized by the number of rays $N$ in this sampling process. 

\begin{figure}[!t]
    \centering
    \includegraphics[width=\textwidth]{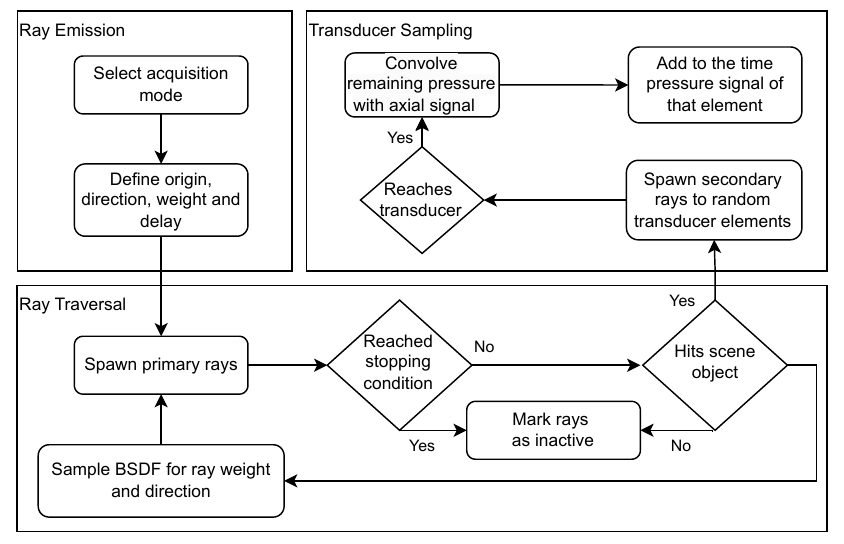}
    \caption{Flow diagram on the main parts of the ray tracing: the ray emission based on the transducer geometry, the ray traversal through the scene, and the transducer sampling to enhance the likelihood of the rays reaching the transducer}
    \label{fig:flow_diagram_ray_tracing}
\end{figure}

Similarly, we can express the general rendering equation in the context of Monte Carlo integration as:

\begin{equation} \label{eq:mc_rays_general_equation}
P_o(\mathbf{x}, \omega_o) \approx P_e(\mathbf{x}, \omega_o)+ \frac{1}{N} \sum_{i=1}^N \frac{f_d(\mathbf{x}, \omega_i, \omega_o) P_i(\mathbf{x}, \omega_i) (\mathbf{n} \cdot \omega_i)}{p(\omega_i)}
\end{equation}
where the integral over the hemisphere is again replaced by a random sampling strategy. Each sampled direction $\omega_i$ contributes to the result, weighted by the inverse of its probability density $p(\omega_i)$, which accounts for the likelihood of sampling this direction based on the geometrical terms and the BSDF. The result is further normalized by the total number of ray $N$ used in this sampling process. In the following, we describe the mechanisms that take place at every step in the simulation and mention the formulas that are based on the underlying physics (See figure \ref{fig:flow_diagram_ray_tracing} for further clarification).

\subsection{Proposed approach} \label{sec:proposed_approach}

\subsubsection{Ray Emission}
\begin{figure}[ht]
    \centering
    \includegraphics[width=0.95\textwidth]{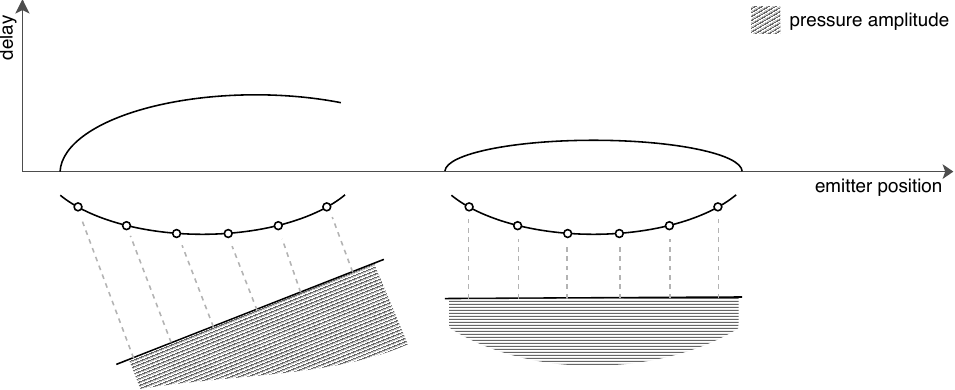}
    \caption{Visualization of two plane wave image transmission event with different angles. The added delays depend on the transmission angle and the transducer geometry. Due to the cosine weighting described in equation \ref{eq:f_d}, we see the change in the amplitude of the pressure over the rays.}
    \label{fig:plane_wave_visualization}
\end{figure}

In standard ultrasound acquisition, pressure waves are emitted from the transducer elements into the scene. As the ultrasound pressure wave is not continuous but is normally limited to several wavelengths, specific pressure fields are formed based on constructive or destructive interference of the fields. Special acquisition strategies use this phenomenon to create focused, diverging, or planar wavefronts by delaying the emission of the elements in the scene. We adopt a similar strategy for plane wave imaging scenario in the proposed method. As shown in figure \ref{fig:plane_wave_visualization}, we are adding delays based on the transmission angle. Due to the independence of the rays from each other, no pressure overlay or interference is happening during the ray tracing - contrary to classic wave simulators. We can, therefore, emit rays with different angles and delays, belonging to different "acquisition events" at the same time.
In order to simulate directivity of the transducer elements, we are adding a weighting function $f_d$ to the outgoing pressure $P_o$ in direction $\omega_o$ with normal $\mathbf{n}$. The outgoing pressure is normalized by the number of rays $N$ over the whole transducer surface. 

\begin{equation}
    f_d(\omega_o) =\frac{1}{N} (\omega_o \cdot \mathbf{n})
    \label{eq:f_d}
\end{equation}

In equation~\ref{eq:f_d}, $\cdot$ symbolizes the dot product and, therefore, represents a cosine weighting function. We define the transducer based on the number of lateral elements, the elevational extent, the radius, and the opening angle. Similarly, we define the number of rays to sample on that transducer element. We are randomly distributing the rays on the emitter surface and, in the case of plane wave imaging, randomly selecting an angle depending on the chosen acquisition scheme.

\subsubsection{Ray Traversal}
In the following, we refer to the rays that reflect and refract through the scene as primary rays. Rays that are cast from a specific interaction point towards a selected target point are considered secondary rays. After emitting a primary ray into the scene, we follow its path until it either fails to intersect any object and becomes inactive or encounters a boundary. In the latter case, we first trigger a transducer sampling strategy (explained in the next paragraph) and then evaluate the bidirectional scattering distribution function (BSDF) at the intersection point to determine the new direction and amplitude of the resulting ray. 

Similar to previous work \cite{mattausch2018realistic,amadou2024cardiac}, reflection and transmission at tissue boundaries are governed by the acoustic impedance ratio $\eta = Z_1 / Z_2$. Using Snell’s law, we derive the directions of the reflected $\omega_r$ and transmitted $\omega_t$ rays from the incident direction $\omega_i$ and the surface normal $\mathbf{n}$: 
\begin{equation}
    \cos\theta_r = \mathbf{n} \cdot (-\omega_i)
\end{equation}
\begin{equation}
    \cos\theta_t = \sqrt{1 - \eta^2 (1 - \cos^2\theta_r)}
\end{equation}
\begin{equation}
    \omega_r = \omega_i + 2\cos \theta_r\mathbf{n}
\end{equation}
\begin{equation}
    \omega_t = \eta\omega_i + (\eta \cos\theta_r-\cos\theta_t)\mathbf{n}
\end{equation}
Here, $\theta_r$ and $\theta_t$ are the reflection and transmission angles, respectively. The amplitude of the reflected wave can be calculated based on the Fresnel equation: 
\begin{equation}
    A_r = \frac{Z_1 \cos \theta_r - Z_2 \cos \theta_t}{Z_1 \cos \theta_r + Z_2 \cos \theta_t} 
\end{equation}

and the intensity of the reflected intensity is  $I_r = I_i * (A_r)^2$. Due to energy conservation, the following equation must hold: $I_i = I_t +I_r$. In the Monte Carlo sampling scheme, we have to decide whether the ray reflects or transmits. In order to maintain a physically consistent reflected-to-transmitted ratio, we sample a variable $y$ from a uniform distribution $y\sim\mathcal{U}(0,1)$ and reflect if $y < (A_r)^2$.

Contrary to existing works, we employ a microfacet distribution to account for surface roughness, enabling the modeling of both specular and diffuse scattering. Rather than representing the surface as a single planar entity with a uniform normal direction, a microfacet distribution models it as a collection of numerous tiny planar elements (microfacets), each with a normal direction that deviates according to the specified distribution. The GGX microfacet distribution \cite{walter2007microfacet}, a commonly used microfacet distribution model in physics-based rendering \cite{nimier2019mitsuba}, provides a PDF for the orientations of these microfacets given by:
\begin{equation}
D(\mathbf{h}) = \frac{\alpha^2}{\pi\bigl((\mathbf{n}\cdot\mathbf{h})^2(\alpha^2-1)+1\bigr)^2},
\end{equation}
where $\alpha$ is the roughness parameter, $\mathbf{h} = (\omega_i + \omega_o)/\|\omega_i+\omega_o\|$ is the half-vector, and $\mathbf{n}$ is the surface normal. Lower $\alpha$ values concentrate orientations near the normal, producing predominantly specular reflections, while higher $\alpha$ values lead to more diffuse scattering. Incorporating this GGX-based model into the proposed acoustic ray tracing framework allows a seamless transition from smooth, specular surfaces to rough, diffuse ones, offering a physically accurate representation of wave interactions at tissue boundaries. In the context of equation \ref{eq:mc_rays_general_equation}, the GGX-based microfacet distribution directly influences the sampling strategy by defining the probability $p(\omega_i)$ and the BSDF $f_d$, which together determine the weighting and directionality of scattered rays. In order to limit the ray tracing we allow a maximum number of interactions with the scene.

\subsubsection{Transducer Sampling}

To enhance the likelihood of rays reaching the transducer, we use a targeted ray tracing approach. In this step, we are casting secondary rays from the interaction points towards a specified point on the surface of the transducer. This point is randomly sampled from the center points of the transducer elements.
These secondary rays interact with the scene, accumulating pressure reductions similar to the primary rays. If a secondary ray encounters a boundary, the corresponding BSDF is evaluated. Additionally, the probability of continuing along the original direction toward the selected transducer element point is incorporated into further calculations. This approach essentially repeats the process described in the previous paragraph but restricts it to a single predefined direction toward the target transducer element.
To account for the directivity of the elements in the transducer, we are adding a weighting function $f_d$, which, for each ray received, computes the contribution based on the incoming direction $\omega_i$ and the normal $n$ of that transducer element (similarly to the emitting case). We are defining two angles in order to weigh the contribution to the transducer element: a main beam angle $\alpha_m$ and a cutoff angle $\alpha_c$:  
\begin{equation}\label{eq:cutoff_angle}
f_d(\omega_i) = 
\begin{cases} 
0, & \text{if } |\alpha| > \alpha_c, \\
\frac{\alpha_c - |\alpha|}{\alpha_c - \alpha_m}, & \text{if } \alpha_m < |\alpha| \leq \alpha_c, \\
1, & \text{if } |\alpha| \leq \alpha_m.
\end{cases}
\end{equation}

where $\alpha$ is the angle between the incoming direction $\omega_i$ and the transducer element normal $\mathbf{n}$, computed as: $\alpha = \arccos(\mathbf{n} \cdot \omega_i)$. This can consequently be integrated into equation \ref{eq:mc_emitter}.

\subsubsection{Phase Calculation and Signal Processing}
When a ray successfully reaches the transducer surface, its remaining pressure value is recorded into the time-pressure signal of the corresponding transducer element. This signal is then convolved with a sinusoidal function windowed by a Gaussian envelope. This operation effectively broadens the pressure signal and introduces phase information, mimicking the axial pulse response of a point spread function.

The resulting axial signal $s(t)$ at time $t$ is expressed as:

\begin{equation} \label{eq:ax_signal} s(t) = \sin(2 \pi f_c t) \exp\left(-\frac{t^2}{\sigma}\right), \end{equation} 

In Equation~\ref{eq:ax_signal}, \( f_c \) denotes the central frequency of the transducer, while \( \sigma \) represents the standard deviation of the Gaussian windowing function. The value of \( \sigma \) is determined based on the number of wave cycles emitted by the transducer. Once the contributions from all rays are incorporated into the time-pressure signals of the individual transducer elements, the ray tracing process is complete. The acquired data is then processed through a standard signal processing pipeline. Specifically, we apply a classic delay-and-sum beamforming method for the plane wave data, followed by demodulation and log compression. Finally, by limiting the dynamic range, the data is converted into the final B-mode image.

\section{Implementation details}

The implementation of the ray tracing is built on top of the Mitsuba 3 software \cite{wenzel2022mitsuba}, a physics-based renderer for forward and inverse light transport simulation of natural images. Mitsuba is written in C++ with Python bindings and offers flexibility through its modular framework. Additionally, the light transport equations are fully differentiable, enabling inverse transport simulations. The software can be built for various renderer variants that run on either CPU or GPU. For this work, we use the GPU variant, which leverages NVIDIA's OptiX rendering framework \cite{parker2010optix}.

To adapt Mitsuba for ultrasound simulation, we derived custom classes in Python and C++ for the emitter, sensor, memory block, reconstruction filter, film, and integrator. Further details about the Mitsuba framework can be found in its documentation\footnote{\url{https://mitsuba.readthedocs.io/en/stable/}}. 

For beamforming and signal processing, we utilize the Ultraspy library \cite{ecarlat2023ultraspy}. The ray tracing simulations are executed on a desktop PC equipped with an Intel i7-7200 processor (20 cores) and an NVIDIA RTX 4070 Ti GPU.

\section{Experiments}

We compare the proposed simulator with the simulator from \cite{mattausch2018realistic} due to its open-source availability and comparable ray tracing capabilities. From now on, we refer to it as the baseline simulator. Two images are generated to evaluate and compare the performance of each simulator, with additional comparisons made to a real ultrasound acquisition.

We are evaluating images acquired from a spine phantom. We, therefore, scan longitudinally and transversely one position on the phantom. To acquire the sample images with the real ultrasound machine, the spine phantom is submerged underwater, and a specific vertebra is selected for scanning. The acquisition is performed with a Siemens Juniper Acuson system equipped with a convex probe (5C2). This probe has a center frequency of 5 MHz, 128 transducer elements, and an opening angle of 70 degrees.

To align the real acquisition with the synthetically created ones, we perform a CT scan of the phantom. In order to load it into the simulators, we are converting the obtained scan to a mesh using the software ImFusion\footnote{https://www.imfusion.com/}. The mesh consequently is positioned in Blender\footnote{https://www.blender.org/}, aligning the virtual transducer with the real setup. Despite these efforts, small positioning errors may still be present, limiting the comparison to a qualitative evaluation of the simulated and acquired B-mode images.

In the proposed simulator, specific parameters must be configured for ray tracing and beamforming. For imaging, we adopt a plane wave imaging scheme, acquiring 25 plane waves with angles ranging from [-30\textdegree, 30\textdegree]. The acquisition setup tries to closely match the real acquisition, including a central frequency of 5 MHz, 128 transducer elements, and a dynamic range of 90 dB. Additionally, we set the sampling frequency to 50 MHz, the pulse duration to 5 cycles, and an approximated elevational beam width of 4 mm, consistent with similar transducers from other vendors \cite{scholten2023differences}. The ray tracing process is constrained to a maximum path length of 20 cm, with a limit of 10 bounces per ray before termination. Additionally, each transducer element emits 100,000 rays, resulting in approximately 13 million rays emitted into the scene simultaneously. Ray tracing is typically completed within 1 second. As long as the possibility is given, we set the same parameters in the baseline simulator (e.g., setting the central frequency to 5 MHz). For optimal results, we set the cutoff angle and the beamwidth angle of the directivity function defined in equation \ref{eq:cutoff_angle} to 2\textdegree.  

For simulating the vertebra, we use the bone presets in the baseline simulator \cite{mattausch2018realistic}. For the surrounding water, we set the scattering parameters close to zero and set the acoustic impedance to 1.54. The impedance values from the baseline are adapted into UltraRay (bone: 7.8 and water: 1.54). Additionally, we empirically determined a surface roughness value of 0.5 for the BSDF of the bone, yielding the most realistic results.

\begin{figure}[!t]
    \centering
    \includegraphics[width=0.9\textwidth]{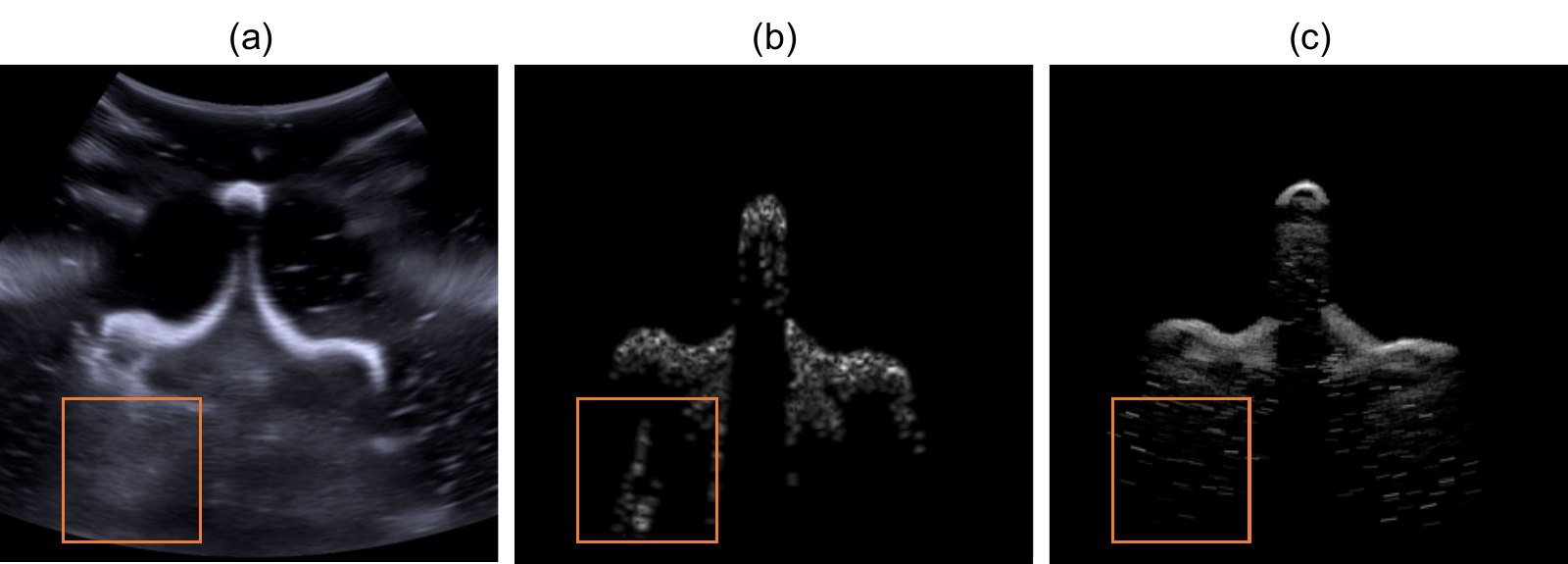}
    \caption{Comparison of (a) real acquisition, (b) baseline simulator, and (c) UltraRay. The orange box highlights strong unrealistic reflections observed in the baseline simulator, which are absent in both the real acquisition and UltraRay.}
    \label{fig:comparison_acquisition}
\end{figure}

When comparing the results in Figure~\ref{fig:comparison_acquisition}, the real acquisition is noticeably more affected by noise and artifacts, largely due to reverberations and reflections in the water bath environment. Both the baseline simulator and UltraRay successfully capture the general geometry of the vertebra. However, when examining the shape outline and intensity, the proposed approach demonstrates a closer alignment with the real acquisition. The real B-mode image also reveals a thicker bone structure, likely attributed to strong attenuation and backscattering within the bone. Additionally, reflections in the real B-mode image are observed to be more pronounced at locations where the bone surface is parallel to the transducer surface, a phenomenon that is similarly captured in our simulations. However, the brightness reduction present in UltraRay is less pronounced in the real acquisition.

\begin{figure}[htb]
    \centering
    \includegraphics[width=0.9\textwidth]{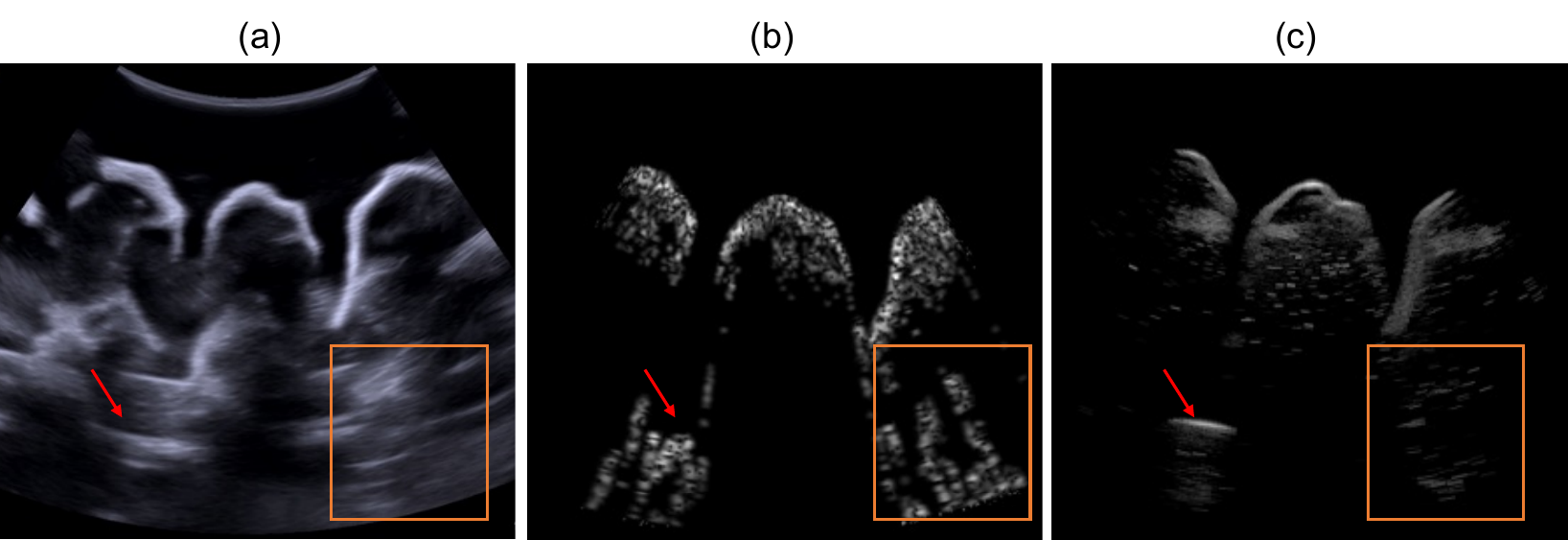}
    \caption{Comparison of (a) real acquisition, (b) baseline simulator, and (c) UltraRay. The orange box highlights unrealistic reflections present in the baseline simulator compared to the real and the proposed method. Additionally, the red arrow showcases a realistic reflection in UltraRay that is also present in the real acquisition while it appears highly distorted in the baseline. }
    \label{fig:comparison_acquisition_rotated}
\end{figure}

The same observations can be made when analyzing the longitudinal scan (Figure~\ref{fig:comparison_acquisition_rotated}). In this case, the baseline simulator again produces a realistic first interaction line. However, unrealistic reflections can be observed within the regions highlighted by the orange box and indicated by the red arrow. These artifacts arise because the baseline simulator considers an echo directly without accounting for the return path. This results in an accumulation of signals in areas where rays become entrapped, leading to unrealistic artifacts in the B-mode image. Such artifacts are absent in both the real acquisition and the proposed simulator.

\section{Discussion and Conclusion}

The presented simulator lays a robust foundation for physics-based ultrasound simulation using differentiable ray tracing. At its current stage, the simulator effectively models surface interactions at tissue boundaries, and can be extended to include other interactions. Future enhancements could focus on incorporating scattering and attenuation, inspired by existing ray tracers in ultrasound.

Currently, we restrict the opening angle of transducer elements during the transducer sampling strategy to enhance image quality. Exploring methods to manage noise without sacrificing a wider opening angle would further improve the simulator’s versatility and realism, making it more applicable to a broader range of ultrasound imaging scenarios.

We introduced a method for incorporating phase information based on the distance traveled, laying the groundwork for accurately modeling the phase received at the transducer. While effective, variations in the speed of sound and transitions between different tissue types can distort phase information. To address this, a promising direction for improvement could involve adapting Mitsuba 3’s internal mechanism for tracking light polarization to ultrasound. 

In conclusion, we presented a novel ray tracer built upon a differentiable, physically based rendering framework. The proposed approach traces rays from the transducer through the scene and back to the emitter, offering a detailed simulation of ultrasound imaging. Following a beamforming strategy, we define a ray emission scheme, include a signal processing pipeline, and show the efficacy of UltraRay by demonstrating plane wave imaging. We derived and introduced the foundational equations underlying this simulation. Additionally we showcase the benefits of UltraRay in comparison to state-of-the-art ray tracing simulators and demonstrate the enhanced realism of simulated images using UltraRay. With this work, we aim to establish a solid foundation for future advancements in ultrasound simulation using physics-based differentiable ray tracing.

%
%
%
%
\bibliographystyle{splncs04}
\bibliography{ultraray.bib}
\end{document}